\documentclass[]{article}
\usepackage{natbib}
\usepackage[a4paper,left=3cm,right=3cm,top=3cm,bottom=3cm]{geometry}
\usepackage{lmodern}
\usepackage{amssymb,amsmath}
\usepackage{ifxetex,ifluatex}
\usepackage{fixltx2e} 
\ifnum 0\ifxetex 1\fi\ifluatex 1\fi=0 
  \usepackage[T1]{fontenc}
  \usepackage[utf8]{inputenc}
  \usepackage{textcomp} 
\else 
  \usepackage{unicode-math}
  \defaultfontfeatures{Ligatures=TeX,Scale=MatchLowercase}
\fi
\IfFileExists{upquote.sty}{\usepackage{upquote}}{}
\IfFileExists{microtype.sty}{%
\usepackage[]{microtype}
\UseMicrotypeSet[protrusion]{basicmath} 
}{}
\IfFileExists{parskip.sty}{%
\usepackage{parskip}
}{
\setlength{\parindent}{0pt}
\setlength{\parskip}{6pt plus 2pt minus 1pt}
}
\usepackage{hyperref}
\hypersetup{pdfborderstyle={/S/U/W 1},
            breaklinks=true}
\usepackage{longtable,booktabs}
\IfFileExists{footnote.sty}{\usepackage{footnote}\makesavenoteenv{longtable}}{}
\usepackage{graphicx,grffile}
\makeatletter
\def\maxwidth{\ifdim\Gin@nat@width>\linewidth\linewidth\else\Gin@nat@width\fi}
\def\maxheight{\ifdim\Gin@nat@height>\textheight\textheight\else\Gin@nat@height\fi}
\makeatother
\setkeys{Gin}{width=\maxwidth,height=\maxheight,keepaspectratio}
\ifx\paragraph\undefined\else
\let\oldparagraph\paragraph
\renewcommand{\paragraph}[1]{\oldparagraph{#1}\mbox{}}
\fi
\ifx\subparagraph\undefined\else
\let\oldsubparagraph\subparagraph
\renewcommand{\subparagraph}[1]{\oldsubparagraph{#1}\mbox{}}
\fi

\makeatletter
\def\fps@figure{htbp}
\makeatother

\title{\textsc{Mitigating Language Barriers in~Education: Developing Multilingual Digital Learning Materials with Machine Translation}}

\date{}
\author{Lucie Poláková, Martin Popel, Věra Kloudová, Michal Novák, \\Mariia Anisimova, Jiří
Balhar\\\textit{Charles University, Faculty of Mathematics and Physics\footnote{\texttt{\{surname, mnovak\}@ufal.mff.cuni.cz; balhar.j@gmail.com}. Address: Institute of Formal and Applied \mbox{Linguistics}, MFF UK, Prague, Czech Republic}
}} 

\begin{document}
\maketitle

\begin{abstract}
\noindent{The EdUKate project combines digital education, linguistics, translation
studies, and machine translation to develop multilingual learning
materials for Czech primary and secondary schools. Launched through
collaboration between a major Czech academic institution and the
country's largest educational publisher, the project is aimed at
translating up to 9,000 multimodal interactive exercises from Czech into
Ukrainian, English, and German for an educational web portal. It
emphasizes the development and evaluation of a direct Czech--Ukrainian
machine translation system tailored to the educational domain, with
special attention to processing formatted content such as XML and PDF
and handling technical and scientific terminology. We present findings
from an initial survey of Czech teachers regarding the needs of
non-Czech-speaking students and describe the system's evaluation and
implementation on the web portal. All resulting applications are freely
available to students, educators, and researchers.\\

}
\end{abstract}

\section{Introduction}
\label{introduction}

In today's rapidly evolving educational landscape, digital learning
helps make education more accessible and inclusive for all students. It
provides flexible, scalable tools for teaching and learning, making it
valuable for diverse and multilingual classrooms.

The \emph{EdUKate} project (Promoting Digital Education of
Foreign-Language Children through Machine Translation) is based on a
close collaboration between \href{https://ufal.mff.cuni.cz}{{The
Institute of Formal and Applied Linguistics}}, Faculty of Mathematics
and Physics, Charles University, and \href{https://www.fraus.com}{{Fraus
Publishing}}, the largest Czech textbook publisher. In this project, we
focus on increasing the reach of digital education within the Czech
school system, particularly for non-native speakers of Czech, including
both children and their parents. Our approach centers on the development
of multilingual interactive content, namely educational materials from
the Czech learning portal
\href{https://www.skolasnadhledem.cz/}{\emph{{Škola s nadhledem}}}
\citep{fraus_skolasnadhledem}, which we translate into multiple languages to reach a broader
user base: These educational resources are intended not only for
non-native speakers but also for all students in primary and secondary
schools across the Czech Republic. For the translation of these
materials, we employ specially adapted machine translation (MT)
techniques.

The choice of target languages is informed primarily by political
developments in Europe over the past three years, which have resulted in
an increased number of individuals living involuntarily outside their
home countries. In the Czech Republic, this shift has led to a
significant population of Ukrainian children whose access to education
depends on their successful integration into a foreign society and
educational system. These students need access to quality education in
their mother tongue, support in learning Czech and foreign languages
based on their native language, and in learning various subjects in
Czech. Therefore, selected learning materials have already been
translated into Ukrainian. Based on further analysis of the situation of
multilingual students in Czech schools, we have decided to expand our
translation efforts to include English and German.

In this article, we provide insight into how educational materials in
foreign languages can be effectively made accessible to children
with different mother tongues. We outline the key challenges encountered
in the process, as well as the range of linguistic, pedagogical, and
technical considerations involved in developing such resources. The
following sections provide an overview of our project: In the
methodology section (2), we briefly present the data used, the MT
techniques applied, and the methods used to assess the situation of the
target user group. The subsequent section (3) outlines the main findings
of the project, including the results of both automatic and manual
evaluation, as well as a linguistic and translation-theoretical
analysis. We also provide a sample of one of the original exercises
alongside its translated version. The final section (4) discusses the
conclusions of our findings and offers reflections on how our current
approach could be further improved in the future.

\section{Methodology}\label{methodology}

\subsection{Data: Škola s nadhledem}
\label{skola-s-nadhledem}

\emph{Škola s nadhledem} is an educational web portal created as an
initiative of the Fraus Publishing House. Its goal is to help children
develop their knowledge through interactive online or hybrid practice.
In more than 9,000 online exercises, it covers school curriculum from
primary to high school (i.e.~the content targets approx. 6 to 19 years
of age) and across a wide range of school subjects. The interactive and,
to a large extent, multimodal content includes drag-and-drop exercises,
yes/no and multiple-choice quizzes, fill-in-the-blank tasks, matching
pairs, crosswords, mind maps, dictation exercises, etc. In the last
three years, the portal has had more than 1.5 million users and more
than 100,000 practicing students every month. The web portal is also
part of the concept of hybrid educational materials -- a unique
combination of printed textbooks or workbooks and interactive practice
with immediate feedback. The users, regardless of whether or not they
currently use Fraus printed textbooks at school, have the opportunity to
practice topics of their choice for free.

From the perspective of creating a multilingual mutation of the portal
by translating the original exercises to other languages, it should be
noted that not all available exercises make sense to translate. For
instance, image-based or syllable-based exercises for first graders,
crosswords, and grammar practice would not preserve their original
purpose. Such exercises were excluded from the data we work with.
Further, translation-wise, the web portal data is quite specific: the
exercises often consist of very short texts with rare occurrence of
multi-sentence context, there is a lot of subject-specific, professional
terms, and, last but not least, the content is html-formatted, including
many images or other objects. All these factors make the task
challenging even for the best available MT systems.

Established methodology of MT quality evaluation requires reference
data, i.e. ideally a human-translated subset of the source-language
texts serving for comparison with the MT outputs. We acquired such an
evaluation dataset for Czech--Ukrainian through manual translation of
selected exercises from \emph{Škola s nadhledem}. We selected 396
exercises of various types in the subjects of biology, chemistry, and
geography. These were divided into a development set (190 exercises) and
a test set (206 exercises). Both datasets were manually translated from
Czech into Ukrainian by professional translators. The same process is
planned for English and German. For further insight on data processing,
see~\ref{machine-translation-methods}.

\hypertarget{methods}{%
\subsection{Methods}\label{methods}}

Our methodology includes an initial survey of the language situation in
the target user group, MT methods, and a variety of evaluation methods.
These are described in detail in the following subsections~\ref{monitoring}--\ref{evaluation-methods}, respectively, and the results of the evaluations are presented in
Section~\ref{results}.

\hypertarget{monitoring-the-needs-of-the-target-group}{%
\subsubsection{Monitoring the needs of the target group}
\label{monitoring}}

In the initial phase of monitoring the needs of the target group, we
analyzed the 2023 report \emph{Education of Refugee Children in the
Czech Republic}, prepared by the PAQ Research agency in collaboration
with the Institute of Sociology of the Czech Academy of Sciences
\citep{safarova2023vzdelavani}. Our focus was on the situation of Ukrainian refugees in
relation to children's education, their leisure activities, and their
knowledge of the Czech language. The research was based on interviews
with 751 households, comprising 1,203 children aged 3--17 years. One of
the key findings was that participation in education had slightly
increased compared to the previous monitoring period. However, it was
also revealed that children aged 6--17 had limited or no access to a
computer in 47\% of households, which hindered their ability to
participate in education, including remote learning. The report
therefore identified increasing educational participation and providing
support to Ukrainian pupils as key priorities.

As part of our monitoring of the target group needs, we conducted our
own questionnaire survey among primary and secondary school teachers.
The respondents were teachers piloting educational materials in
cooperation with Fraus Publishing, as well as from the professional
networks of team members and teacher communities on social media. We
received responses from 42 participants, 31 of whom teach at the lower
or upper secondary level, and 11 at the primary level. The survey
results indicated that each respondent taught at least one student whose
native language was Ukrainian.

The key findings that served as a basis for the next steps in our
project can be summarized as follows: Most teachers reported that pupils
with a different mother tongue do not have access to a teaching
assistant during lessons (69\%). At the same time, however, 83\% of
respondents stated that their schools offer pupils the opportunity to
attend individual Czech language lessons for foreigners. According to
the survey, though, 71\% of pupils attend Czech lessons instead of a
different school subject. Furthermore, when a pupil with a different
mother tongue does not understand the instruction in general, most
teachers try to explain it more clearly in Czech. As a secondary
strategy, many also reported using online translation tools or involving
another pupil who shares the same mother tongue.

In terms of the challenges faced by pupils, the most commonly reported
obstacles --- besides the rules of Czech grammar -- include the general
language barrier, subject-specific terminology (especially in natural
sciences), and the pace of instruction, which can be difficult to follow
due to limited comprehension. At the same time, preparing individualized
instruction is time-consuming, and the textbooks used in class offer
little support in this regard.

We also explored the extent to which online learning materials /
practice exercises are used in teaching. 76\% of respondents stated that
they have access to the internet in every lesson, and further 17\%
reported that they can use the internet regularly. Regarding the
specific use of online resources, most teachers reported using
interactive whiteboards, followed by tablets, mobile devices, or a
combination of these tools. When asked whether they use online practice
platforms during instruction, 61\% of respondents answered positively.
Among the specific platforms mentioned, the Czech educational portal
\emph{Škola s nadhledem} was cited several times.

At the end of the questionnaire, we asked whether teachers and their
students would benefit from the option to switch online practice
exercises into the native language of pupils with different mother
tongues. 40\% of respondents indicated that this feature would be highly
helpful, and another 44\% reported that they would make use of it in at
least some cases.

\hypertarget{machine-translation-methods}{%
\subsubsection{Machine translation methods}
\label{machine-translation-methods}}

In spring 2022, with the help of many language data providers, we
developed the first version of an automatic translator between Ukrainian
and Czech under the name Charles Translator (CHT) \citep{popel2024edukate,popel2024charles}. This was a
quick way to meet the demand for such a service, which at that time was
not available in the required quality even from commercial companies.
The system uses technologies developed earlier in the English--Czech
neural translator CUBBITT \citep{popel2020transforming}, which unlike others applies the Block
Backtranslation method. This allows for efficient use of monolingual
training data. The components of the translator currently include the
translation service and multiple interfaces to access it: a web
interface (\url{https://translator.cuni.cz/}), Android app, and API.
The API is the most relevant interface to our project, as a version of
it was tailored specifically for the needs of Fraus Publishing.

Unlike the common commercial translators that pivot the translation over
English, Charles Translator translates directly and is thus not prone to
some types of errors, especially in certain grammatical categories. The
Charles Translator license allows free use via API for non-commercial
purposes, thus it is possible to include this translator in educational
activities that will be free of charge for the users.

\textbf{Domain adaptation}

Translating texts from the educational domain presents specific
challenges that general-purpose MT models often fail to address
effectively. To ensure higher translation quality for domain-specific
content, it is necessary to adapt translation models to the
characteristics of the given domain. The main advantage of having our
own translator is that we can adapt it to translate texts typical of
educational materials (domain adaptation), which would not be possible
if we used a third-party MT service. This adaptation itself relies on
the availability of substantial amounts of in-domain data, which are
crucial for both training and evaluation. For training, large volumes of
data are desirable, as modern neural MT systems can learn effectively
even from lower-quality (noisy) datasets. In contrast, evaluation data
must be of the highest quality, though only a limited amount is
required, as it serves to assess the performance of the translation
system.

In addition to domain adaptation, several new functionalities of the
translator are being gradually developed. The most important new feature
is support for translating formatted text, such as the HTML or XML
format used in the \emph{Škola s nadhledem} database. It relies on a
word alignment tool that determines where to insert formatting tags
(e.g. bold or italics markers) within the translated sentences. This
functionality is built into the API server.

\hypertarget{evaluation-methods}{%
\subsubsection{Evaluation methods}\label{evaluation-methods}}

A machine translation project of our type involves different kinds and
stages of evaluation. Part of the evaluation depends on the development
of a human-translated test set (see~\ref{skola-s-nadhledem} above), further we use automatic
(\ref{automatic}) and human (\ref{human}) MT evaluation methods, and finally, we performed a
pilot test of the translated content by selected users (\ref{pilot}).

\hypertarget{results}{%
\section{Results}\label{results}}

\subsection{Automatic evaluation of machine translation}
\label{automatic}

For the automatic evaluation of the system performance, we have
translated the development set part of the data, which consists of 3,770
sentence-like segments (approx. 21,000 words). We used the version of
Charles Translator from 2023 as a baseline. Table 1 shows the
performance of the baseline and the best adapted system (Adapted CHT
2024), which was developed using translationese tuning {[}5{]} and
additional monolingual and parallel training data. This system showed
improvements in both the automatic chrF score \citep{popovic2015chrf} and human
evaluation, and was therefore chosen to translate the first set of
exercises for the pilot testing of the portal (see~\ref{pilot} below).
\\
\begin{longtable}[]{@{}lll@{}}
\toprule
\endhead
\emph{\textbf{System/Metrics}} & \emph{\textbf{Automatic (chrF)}} &
\emph{\textbf{Human (0-10)}}\tabularnewline
Charles Translator 2023 & 61.6 ± 0.8 & 7.41 ± 3.35\tabularnewline
Adapted CHT 2024 & \textbf{63.1 ± 0.8} & \textbf{7.80 ±
3.25}\tabularnewline
\bottomrule
\\
\caption{Automatic and human evaluation of translation quality}
\label{tab_scores}
\end{longtable}

\subsection{Human evaluation}
\label{human}

For the quantitative human evaluation of translation quality, we have
selected a subset of 1,600 segments, which were distributed among 7
annotators, native speakers of Ukrainian with a good knowledge of Czech.
The annotators were asked to assign a score ranging from 0 (completely
wrong translation) to 10 (excellent translation) to each of the
translations produced by three anonymized systems. In addition to the
Czech source text, the annotators were shown also the reference (human)
translation. The average score of the best-performing system was 7.80,
see Table~\ref{tab_scores}.

\hypertarget{error-analysis}{%
\subsubsection{Error analysis}\label{error-analysis}}

We also conducted a qualitative human evaluation in the form of an error
analysis guided by linguistic and translation theory, focusing on three
linguistic levels: lexical (word meaning), morphological (grammatical
form), and syntactic (sentence structure). Most errors occurred at the
lexical level, particularly with the incorrect translation of terms
across subject areas -- most notably in biology. Lexical errors are
critical, as they often significantly alter meaning and hinder
comprehension. In particular, translating specific terminology can be
challenging even for experienced translators: ``to be able to propose a
term, translators must have acquired a sound knowledge of lexical
morphology, lexicology, sociolinguistics and pragmatics. Besides, some
degree of feasibility of use of the suggested term is required. When
confronted with different alternative terms, to decide between choosing
one possibility or coining a new term is not an easy task.'' \citep{cabre2010terminology}

A striking example would be the mistranslation of specialized biological
terminology, such as ``Hálky na rostlinách vytváří žlabatky'' {[}Galls
on plants are created by gall wasps{]}, which illustrates the challenge
even for professional translators. However, apart from the specialized
terminology the challenge of this sentence lies also in the fact that --
due to the grammatical structure of Czech -- it is (for non-experts in
biology) unclear whether \emph{hálky} (galls) is the subject or the
object, and the same applies to \emph{žlabatky} (gall wasps). Such
instances reveal the need for enriching the training data with external
terminological sources. Morphological errors (e.g., incorrect case,
number, or gender) were less frequent and typically did not impede
understanding. Syntactic errors, mainly involving word order, were rare.

We detected three main reasons for terminology errors that occurred in
our translations.~

(1) A specific term was mistranslated in a given context (e.g.
\emph{přetvařovat se} = \emph{to dissimulate}), but we were able to
detect its correct translations in the training data, yet in other
semantic contexts.

\begin{figure}
\includegraphics[width=5.9in,height=3.59722in]{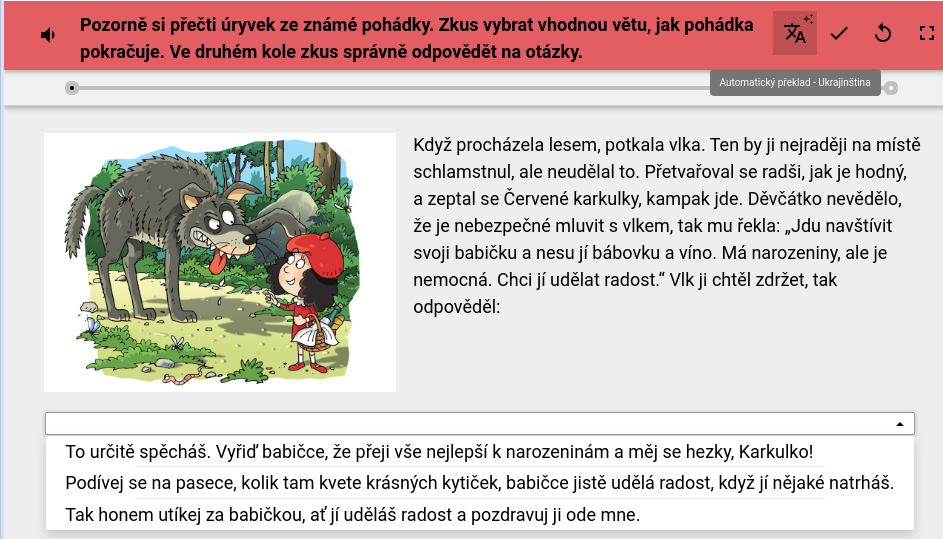}
\label{fig_cs}
\caption{Little Red Riding Hood: a text comprehension exercise
in Czech original}
\end{figure}

\begin{figure}
\includegraphics[width=5.9in,height=3.77778in]{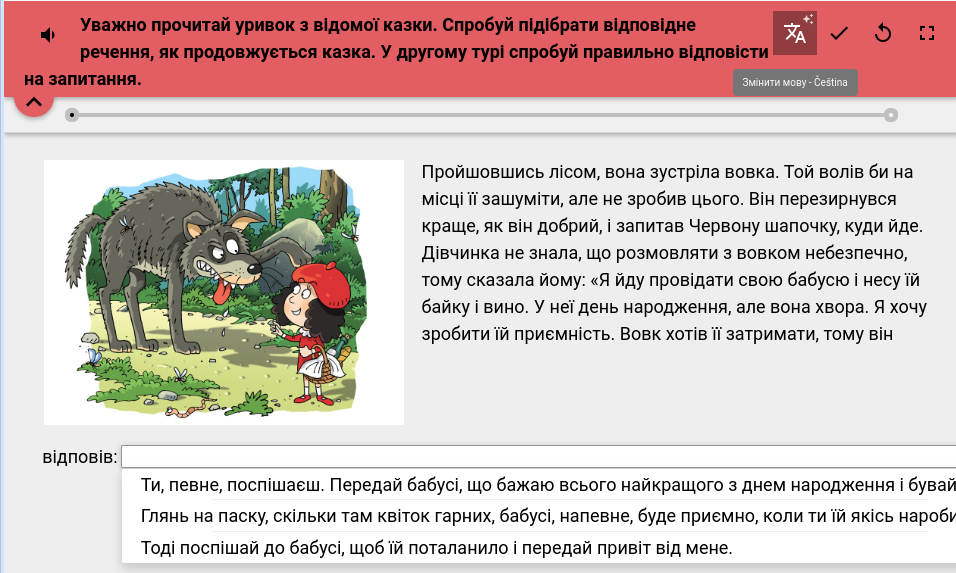}
\label{fig_uk}
\caption{Little Red Riding Hood: a text comprehension exercise
translated to Ukrainian}
\end{figure}

(2) A specific term was mistranslated in any context (e.g.
\emph{schlamstnout} = \emph{to devour} or \emph{bábovka} = \emph{marble
cake}) since there were no occurrences of the term in the training data.
The examples (1) and (2) come from the text comprehension exercise shown
in Fig.~\ref{fig_cs} (Czech) and~\ref{fig_uk} (Ukrainian).

(3) The exact equivalent does not exist at all in the target language.
This situation is represented e.g. by the Czech word \emph{ponravy} in
the following sentence: \emph{Larvy se jmenují ponravy}. {[}The larvae
are called grubs.{]} The word \emph{ponrava} (\emph{ponravy} in plural)
means \emph{a larva of a beetle from the Scarabaeoidea superfamily} and
has no language equivalent in Ukrainian.
\\

\subsection{Pilot testing of the translated portal}
\label{pilot}

The pilot testing of the translated exercises took place in late 2024,
with the aim of validating the project's current outcomes through
qualitative inquiry. The pilot was conducted via structured online
interviews with respondents drawn from among educators and language
professionals. Seven participants were involved in the pilot: primary
and lower-secondary school teachers as well as language experts. All of
them either teach some children whose mother tongue is not Czech or work
in centers that directly support refugee families. The pilot primarily
focused on pupils whose native language was Ukrainian. Participants were
provided with a small set of interactive exercises translated into
Ukrainian, covering various subjects and grade levels. The exercises
were made available on a dedicated website created specifically for the
pilot. The tests were conducted on a selection of exercises translated
by the best performing domain-adapted MT system.~

The outcomes can be summarized into several key areas: (1) the usability
of the translated interactive exercises on the \emph{Škola s nadhledem}
platform, (2) the quality of the translation, (3) the functionality of
the translation. Regarding the usability of the translated interactive
exercises (1), the respondents evaluated the translations as a useful
tool when working with students from non-Czech-speaking backgrounds,
both in school and at home. Respondents agreed that the quality of
translation (2) is of critical importance. In the exercises provided
during the interviews, they identified several issues related to
translation quality: Some translations were difficult to understand in
terms of content, while others contained grammatical errors (see also
our error analysis in Section~\ref{error-analysis}). The respondents emphasized the
importance of precise translations, particularly when it comes to
terminology, which poses the greatest challenge for students. The pilot
testing further revealed that the translated websites were functional
and intuitive (3): respondents understood how to switch between
languages and they found the possibility to toggle between the two
languages during completing the exercises more useful than to display
both languages at once, side by side. The users would welcome the
possibility to display the translation of just a single word within the
other language (tool tip). Some room for improvement was discovered
regarding the responsiveness on different types of devices: computers,
tablets, mobile phones, and interactive whiteboards.

\hypertarget{conclusions}{%
\section{Conclusions}\label{conclusions}}

We presented the EdUKate project, an applied research project focused on
the development and dissemination of multilingual digital learning
materials for primary and secondary school students. The main goal is to
facilitate access to the originally Czech educational web portal
\emph{Škola s nadhledem} for non-Czech-speaking, mainly Ukrainian,
children in the Czech educational system. First, we have conducted a
survey to monitor the educational needs of the target group of users.
Next, we have adapted our own MT methods to be able to translate 1.~between Czech and Ukrainian without pivot; 2.~formatted input (XML, HTML
and other formats), 3.~educational content with more accuracy. We
evaluated the results of the translation using automatic MT metrics and
human evaluation, with the best adapted system scoring an average of
7.80 on a 0‒10 scale in human evaluations. Qualitative linguistic
analysis has shown that the most difficult terms to translate are the
technical ones, while at the same time these are the ones that need to
be translated accurately -- general language is already handled quite
well by most MT systems. We are currently implementing specific methods
(e.g. adding external terminological sources) to eliminate this issue.
We are also experimenting with the latest LLM-based methods to see if
translation quality can be further improved.

Apart from internal MT evaluations, we also cooperate with external
piloting experts. Once a sample of exercises had been translated using
the API, we conducted pilot tests of the portal through interviews with
teachers, who provided valuable feedback on both translation quality and
general technical issues (technical solutions, usability, motivation for
new tools).

In the final, current phase of the project, we are extending support to
English and German, so that by the end of 2026, the portal's content
will be available in four languages. All resulting applications will be
freely available to educators and researchers via the
\href{https://lindat.cz/}{{LINDAT/CLARIAH-CZ}} repository.

\section*{Acknowledgements}

This work was supported by the project TQ01000458 (EdUKate) financed by
the Technology Agency of the Czech Republic
(\href{http://www.tacr.cz}{{www.tacr.cz}}) within the Sigma 3 Programme.
It has been using language resources and tools developed and/or stored
and/or distributed by the LINDAT/CLARIAH-CZ project of the Ministry of
Education, Youth and Sports of the Czech Republic (project LM2023062).

We would like to thank our colleagues at Fraus Publishing: A. Jelínek,
B. Bartošová, K. Berková, M. Sutr, and B. Tvarohová for our ongoing
cooperation and for designing and managing the pilot test.



\bibliographystyle{lrec-coling2024-natbib}
\bibliography{edulearn}

\end{document}